\documentclass{article}

\usepackage{PRIMEarxiv}

\usepackage[utf8]{inputenc} 
\usepackage[T1]{fontenc}    
\usepackage{hyperref}       
\usepackage{url}            
\usepackage{booktabs}       
\usepackage{amsfonts}       
\usepackage{nicefrac}       
\usepackage{microtype}      
\usepackage{lipsum}
\usepackage{fancyhdr}       
\usepackage{graphicx}       
\graphicspath{{media/}}     

\usepackage[numbers]{natbib}
\usepackage{algorithm}
\usepackage{algorithmic}
\usepackage{amsmath}
\usepackage{amssymb}
\usepackage{mathtools}
\usepackage{amsthm}
\usepackage{tikz}
\usetikzlibrary{bayesnet}
\usepackage{multirow}
\usepackage{enumitem}
\title{Fair Inference for Discrete Latent Variable Models
}

\author{
  Rashidul Islam \\
  University of Maryland, \\Baltimore County \\
  \texttt{islam.rashidul@umbc.edu}
   \And
  Shimei Pan \\
  University of Maryland, \\Baltimore County \\
  \texttt{shimei@umbc.edu}
  \And
  James R. Foulds \\
  University of Maryland, \\Baltimore County \\
  \texttt{jfoulds@umbc.edu}
}

\begin{document}
\maketitle

\begin{abstract}
It is now well understood that machine learning models, trained on data without due care, often exhibit unfair and discriminatory behavior against certain populations. Traditional algorithmic fairness research has mainly focused on supervised learning tasks, particularly classification. While fairness in unsupervised learning has received some attention, the literature has primarily addressed fair representation learning of continuous embeddings.  In this paper, we conversely focus on unsupervised learning using probabilistic graphical models with discrete latent variables. We develop a fair stochastic variational inference technique for the discrete latent variables, which is accomplished by including a fairness penalty on the variational distribution that aims to respect the principles of intersectionality, a critical lens on fairness from the legal, social science, and humanities literature, and then optimizing the variational parameters under this penalty. We first show the utility of our method in improving equity and fairness for clustering using na\"ive Bayes and Gaussian mixture models on benchmark datasets. To demonstrate the generality of our approach and its potential for real-world impact, we then develop a special-purpose graphical model for criminal justice risk assessments, and  use our fairness approach to prevent the inferences from encoding unfair societal biases. 
\end{abstract}

\keywords{AI fairness \and Latent variable models \and Variational inference \and Intersectionality}

\section{Introduction}
In our interconnected world, artificial intelligence (AI) and machine learning (ML) have become ubiquitous. Increasingly, the automated decisions made by these systems have important real-life consequences, from credit scoring and subsequent lending outcomes, to college admissions, to career recommendations, to the prediction of re-offending \cite{munoz2016big,o2016weapons}. However, the integrity of these decisions can often be undermined by implicit bias or societal stereotypes in the underlying data, leading the behavior of these learning algorithms to unfairly discriminate against certain groups of people, including women, people of color, and individuals on the LGBTQA spectrum\cite{angwin2016machine,barocas2016big, bolukbasi2016man,munoz2016big,o2016weapons,campolo2017ai,noble2018algorithms}. 

With the rising awareness and regulations, the AI community has devoted much effort to the development and enforcement of numerous quantifiable notions of fairness for  AI/ML models \cite{dwork2012fairness,zemel2013learning,hardt2016equality,kusner2017counterfactual,kearns2018preventing,foulds2020intersectional,keya2021equitable}. The main paradigm for fair algorithms is to posit a mathematical criteria of fairness across protected demographic groups, (e.g. by gender, race, and age) or similar individuals (e.g. persons with similar merits and risks) \cite{berk2018fairness,foulds2020parity}. The paradigm then enforces these criteria, when optimizing objective functions, by penalizing violations \cite{berk2017convex,foulds2020intersectional,islam2021debiasing} or by imposing constraints \cite{zafar2017fairness} or by finding a transformation of data that provides fair latent representations \cite{zemel2013learning,edwards2015censoring,louizos2016variational,madras2018learning,zhao2019inherent,zhao2020conditional}. 

From a fairness perspective, representation learning is appealing because deep learning-based vector representations often generalize to tasks which are unspecified at training time, implying that a properly designed fair network might operate as a kind of ``parity bottleneck,'' reducing discrimination in unknown downstream tasks \cite{creager2019flexibly}. Particularly, the goal of fair representation learning is to transform the data into continuous latent spaces that are invariant to protected attributes and useful to mitigate societal bias in different downstream tasks, e.g., classification \cite{bengio2013representation,locatello2019fairness}. Most of the recent frameworks \cite{louizos2016variational,song2019learning,creager2019flexibly,locatello2019fairness}  are built upon the variational autoencoder (VAE) \cite{kingma2013auto,rezende2014stochastic}, which can perform effective stochastic variational inference (SVI)\cite{hoffman2013stochastic} and learning in continuous latent variable models using backpropagation. 

While the benefits of fair representations in continuous latent space in downstream tasks are clear, we are conversely interested in extending the success of variational techniques to fair inference in graphical models with discrete latent variables. 
As societal prejudices, societal disadvantages, underrepresentation of minorities, intentional prejudices and proxy variables are inherent in historical data \cite{barocas2016big}, inferences can naturally encode these harmful biases in the latent variables which should be prevented to mitigate discriminatory decisions. 
Such graphical models are used in numerous AI/ML methods including 
semi-supervised learning \cite{kingma2014semi}, binary latent attribute models \cite{vahdat2018dvae++}, hard attention models \cite{malinowski2018learning}, clustering \cite{karim2021deep}, and topic modeling \cite{srivastava2017autoencoding}. 
Furthermore, discrete latent variables are a natural fit for complex reasoning, planning and predictive learning, e.g., ``\emph{if you study hard, you will be successful.}'' However, inference on discrete latent variables with backpropagation-based variational methods is difficult because of the inability to re-parameterize gradients \cite{vahdat2018dvae++,van2017neural,shih2020probabilistic,vuffray2020efficient}. To address this, continuous relaxations in the VAE framework have been found effective using the Gumbel-Softmax re-parameterization trick \cite{jang2016categorical,maddison2016concrete} which defines a temperature-based continuous distribution, and converges to a discrete distribution in the zero-temperature limit.  
 \citet{chierichetti2017fair} and \citet{backurs2019scalable} addressed fairness in clustering problems for both k-center and k-median objectives, but not for the general case of graphical models with discrete latents.  
Fairness approaches exist for 
causal latent variable models \cite{kusner2017counterfactual,madras2019fairness}, however, these models enforce fair causal inference in a supervised setting (e.g. with a class label). 

In this paper, we develop a practical framework for fair SVI on arbitrary graphical models with discrete latent variables 
to improve their equity and fairness. 
Our method is general and could be incorporated into probabilistic programming systems without additional assumptions. 
Given a probabilistic graphical model, e.g. a custom model defined for a particular task, our goal is to perform inference such that the results do not reflect negative stereotypes or bias.   
For example, multiple studies demonstrated that police stop people from racial and ethnic minority groups more frequently than whites \cite{harris1999stories,harris1996driving,warren2006driving,gelman2007analysis,carroll2014out}. In a traffic model, we may wish to prevent the inference that individuals of one demographic drive more aggressively than other demographics. Such an inference may in fact be warranted by the data, but it may be known --due to knowledge not encoded in the data, or known causes of the issue such as less data for minorities-- 
that this inference cannot be correct, or it may be desirable to prevent it in order to avoid harm due to the use of the model, pursuant to Title VII \cite{barocas2016big}. Furthermore, our fairness intervention technique enforces  an intersectional fairness notion \cite{foulds2020intersectional} that guarantees fairness protections for all subsets of the protected attributes (e.g., black women, women, and black) which is consistent with the ethical principles of intersectionality theory \cite{crenshaw1989demarginalizing,collins2002black}.

Our key contributions can be summarized as follows:
\begin{itemize}[noitemsep,topsep=0pt]
    \item 
    We develop a method for ensuring intersectional fairness in stochastic variational inference for unsupervised discrete latent variable models. 
    
    \item We demonstrate the utility of our method for clustering analysis using na\"ive Bayes and Gaussian mixture models on fairness benchmark datasets.
    
    \item 
    We show our method's generality by applying it to fair inference in a new special-purpose 
    model for criminal justice 
    decisions 
    based on COMPAS 
    \cite{angwin2016machine}.
\end{itemize}

\section{Background}
\label{background}
\subsection{Problem Setup: Fair Latent Variable Modeling}
We assume a generative model $p(x,z) = p(x|z)p(z)$ that produces a dataset $\mathbf{x}=\{x_{j}\}_{j=1}^{n}$ consisting of $n$ i.i.d. individuals, generated using a set of $K$-dimensional discrete latent variables $\mathbf{z}=\{z_j\}_{j=1}^n$. Let  $A$ be protected attributes, e.g. the individuals' gender and race, which may or may not be included in the typical (or non-sensitive) $D$-dimensional attribute vector $x$. Furthermore, each $z_j$ is assumed to be generated from some prior distribution $p(z)$. 
Variational inference \cite{dayan1995helmholtz,jaakkola1997variational,jordan1999introduction,hoffman2013stochastic} is an optimization approach to solve inference problems for latent variable models. 
We aim to compute the posterior distribution $p(z|x)$, which is assumed intractable over latent variables, and so approximation techniques must be used. The key idea is to approximate $p(z|x)$ with a more tractable distribution $q(z)$, referred to as a variational distribution, and minimize 
the KL-divergence $D_{KL}$ between them.
  
\subsection{Variational Autoencoder (VAE)}
The variational autoencoder (VAE) \cite{kingma2013auto} performs variational inference in a latent Gaussian model where the variational posterior and model likelihood are parameterized by neural nets $\phi$ and $\theta$, respectively. 
\citet{kingma2013auto} developed a differentiable reparameterization technique for efficient SVI for probabilistic models 
in the presence of continuous latent variables $z$. The VAE is generally implemented with a Gaussian prior $p_{\theta}(z) = \mathcal{N}(0,1)$. The objective is to maximize the evidence lower-bound (ELBO): 
\small
\begin{equation}
    \mathcal{L}_{\text{VAE}}(\theta,\phi) = \mathbb{E}_{q_{\phi}(z|x)}[\log p_{\theta}(x|z)] - D_{KL}[q_{\phi}(z|x) || p_{\theta}(z)] \mbox{ .} 
\end{equation}
The objective to maximize $\mathcal{L}_{\text{VAE}}$ is made differentiable by reparameterizing $z\sim q_{\phi}(z|x)$, which is here assumed to be Gaussian, with $z = \mu + \sigma \odot \mathcal{E}$, where $\mathcal{E} \sim \mathcal{N}(0,1)$.    

\subsection{The Gumbel-Softmax Reparameterization Trick}
\citet{jang2016categorical} and \citet{maddison2016concrete} introduced a reparameterization technique for training VAEs with discrete latent variables using the Gumbel-Max trick \cite{gumbel1954statistical,maddison2014sampling}, which provides an efficient way to draw samples $z$ from a $k$-dimensional categorical distribution with class probabilities $\pi$ as $z=\text{one\_hot}({\text{arg max}_i}[g_i + \log \pi_i])$, where $g_i \sim \text{Gumbel}(0,1)$ are i.i.d. samples. As ``$\text{arg max}$" is not differentiable, the softmax function is used 
to approximate it: 
\small
\begin{equation}
\label{eq:gumbel}
    z_i = \frac{\text{exp}((g_i + \log \pi_i)/\tau)}{\sum_{k=1}^{K}\text{exp}((g_k + \log \pi_k)/\tau)} \quad\quad \text{for } i = 1,\dots,K \mbox{ ,}
\end{equation}
where the temperature $\tau$, which controls how closely samples from the Gumbel-Softmax distribution approximate those from the discrete distribution, is annealed towards $0$ during training.

\section{Method: VI with a Fairness Intervention}
\label{sec:method}
We aim to perform fair inference on a generative probabilistic model $p_{\theta}(x,z)$ with discrete latent variables $z$. We desire a stochastic variational inference algorithm to compute the posterior distribution $p_{\theta}(z|x)$ which achieves two properties: 1) inference that scales to the big data, and 2) a simple and efficient fairness intervention using backpropagation. 

\subsection{Inference Network}
Following \citet{kingma2013auto}, we use a neural network-based inference network  $q_{\phi}(z|x)$ for the variational approximation to the intractable posterior $p_{\theta}(z|x)$, where weights and biases of the neural network are variational parameters $\phi$. Following \citet{jang2016categorical}, let the prior $p_{\theta}(z)$ be a discrete distribution with uniform probability $1/K$,  although this can easily be generalized as shown in a later section for special purpose modeling. The first step of our method is to reparameterize the variational distribution for the purposes of sampling latent variables so that discrete distributions are re-represented as unconstrained distributions, in order to facilitate backpropagation on these variables. The two main options are the logistic-normal representation \cite{srivastava2017autoencoding}, and the Gumbel-softmax representation \cite{jang2016categorical, maddison2016concrete}. If we use the logistic-normal representation, we reparameterize $q_{\phi}(z|x)$ using a mean $\mu$ and covariance matrix $\Sigma$ via a logistic function. We focus on the Gumbel-softmax approach, since it performed better in preliminary experiments. To encode discrete $z$, the inference network basically outputs unnormalized log probabilities $\log\pi$ for the latent classes which are then used to reparameterize $q_{\phi}(z|x)$ using the Gumbel-Softmax trick in Equation \ref{eq:gumbel}. 

\subsection{Generative Network}
To learn the generative model's parameters $\theta$, unlike VAE, no neural network is used. It is generally impossible to approximate the true joint distribution over observed and latent variables, including the true prior and posterior distributions over latent variables using VAE framework due to unidentifiability of the model \cite{khemakhem2020variational}. \citet{khemakhem2020variational}, and \citet{zhou2020learning} provided a solution that requires a factorized prior distribution over the latent variables given an auxiliary observed variable, usually class labels. As we desire a framework for unsupervised learning, we present a different approach that also produces identifiable and meaningful latent variables. We randomly initialize $\theta$, while simply considering some hyper-priors $p_{\alpha}(\theta)$ on $\theta$, where $\alpha$ are the hyper-parameters and fixed. In our \emph{vanilla model} with no fairness intervention, we then jointly optimize $\theta$ and $\phi$ via the ELBO: 
\small
\begin{align}
\label{eq:vanilla_elbo}
\mathcal{L}(\theta,\phi) &= \mathbb{E}_{q_{\phi}}[\log p_{\theta}(x,z)] - \mathbb{E}_{q_{\phi}}[\log q_{\phi}(z|x)]\nonumber \\
&= \mathbb{E}_{q_{\phi}}[\log p_{\theta}(x|z)+\log p_{\theta}(z)+p_{\alpha}(\theta)] - \mathbb{E}_{q_{\phi}}[\log q_{\phi}(z|x)]\mbox{ .}
\end{align}

\subsection{Fair Inference Technique}
Our fair inference technique uses fairness criteria from an intersectionality perspective as a penalty term to measure violations, with regard to parity in the inferred discrete latent variables for intersecting protected groups. The inference and learning objective
is: 
\small
\begin{align}
    \label{eq:fair_elbo}
    \min_{\theta,\phi}\; -\sum_{j=1}^{n}\mathcal{L}_{j}(\theta,\phi) + \lambda \mathcal{F}(\phi,A)\mbox{ ,}  
\end{align}
where, $\mathcal{L}(\theta,\phi)$ is the ELBO of the vanilla model in Equation \ref{eq:vanilla_elbo}, $\mathcal{F}$ is a fairness penalty, and $\lambda$ is a hyper-parameter that trades between the ELBO and fairness. To design the fairness penalty on ELBO, we adapt the differential fairness (DF) metric \cite{foulds2020intersectional}, which was originally proposed for classification. 
DF extends the 80\% rule \cite{biddle2006adverse} to multiple protected attributes and 
outcomes, and provides an \emph{intersectionality property}, e.g., if all intersections of gender and race are protected (i.e., Black women), then gender (e.g., women) and race (e.g., Black people) are 
protected:

\emph{Let  $s_1, \ldots, s_p$ be discrete-valued protected attributes, $A = s_1 \times s_2 \times \ldots \times s_p$. An inference mechanism $q_{\phi}(z|x)$ satisfies $\epsilon$-DF with respect to $A$ if for all $x$, and $z \in K$,}
\small
\begin{equation}
	e^{-\epsilon} \leq \frac{p(q_{\phi}(z|x) = z|s_i)}{p(q_{\phi}(z|x) = z|s_j)}\leq e^\epsilon \mbox{ ,} 
	\label{eq:DF}
\end{equation}
\emph{for all   $(s_i, s_j) \in A \times A$ where $P(s_i) > 0$, $P(s_j) > 0$} (Proof given in \cite{foulds2020intersectional}). Smaller $\epsilon$ is better, and $\epsilon= 0$ for perfect fairness, otherwise $\epsilon> 0$. We can measure $\epsilon$-DF using the empirical data distribution. Let $N_{z,s}=\sum_{x \in \mathbf{x}: A = s} \mathbf{z}$ and $N_{s}$ be the empirically estimated expected counts for latent assignments per group and for total population per group, respectively. Then $\epsilon$-$DF$ can be estimated via the posterior predictive distribution of a Dirichlet-multinomial, where scalar $\alpha$ is a Dirichlet prior with concentration parameter $K\alpha$, as:
\small
\begin{equation}
e^{-\epsilon} \leq \frac{N_{z,s_i} + \alpha}{N_{s_i}  + K\alpha}\frac{N_{s_j} + K\alpha}{N_{z,s_j} + \alpha}\leq e^\epsilon \mbox{ .} 
\label{eqn:EDF}
\end{equation}
However, the reliable estimation of $\epsilon$-$DF$ on the inference mechanism in terms of $A$, denoted by $\epsilon(q_{\phi}(z|x),A)$, for a minibatch becomes statistically challenging due to data sparsity of intersectional groups~\cite{foulds2020bayesian}. For example, one or more missing intersectional groups for a minibatch is a typical scenario in the stochastic setting that can lead to inaccurate estimation of the fairness, an obstruction to scaling up the inference using SVI. To address data sparsity in $\epsilon(q_{\phi}(z|x),A)$, inspired by the noisy update technique in several SVI algorithms \cite{hoffman2013stochastic,foulds2013stochastic,islam2019scalable}, we develop a stochastic approximation-based approach that updates count parameters for each minibatch with $m$ datapoints  as follows: 
\small
\begin{align}
    \label{eq:update_count1}
    N_{z,s} &:=(1-\rho_{t})N_{z,s} + \rho_{t}\frac{n}{m}\hat{N}_{z,s}\mbox{ ,}\\
    N_{s} &:=(1-\rho_{t})N_{s} + \rho_{t}\frac{n}{m}\hat{N}_{s}\mbox{ ,}
    \label{eq:update_count2}
\end{align}
where, $\hat{N}_{z,s}$ and $\hat{N}_{s}$ are empirically estimated noisy expected counts per group for a minibatch, and $\rho_{t}$ is a step size schedule, typically annealed towards zero. In practice, we found that fixed $\rho_{t}$, selected as a hyper-parameter, is enough for successfully estimate $\epsilon(q_{\phi}(z|x),A)$ using the global counts $N_{z,s}$ and $N_{s}$, via Equation \ref{eqn:EDF}. The fairness penalty term is then designed as hinge loss: $\mathcal{F}(\phi,A)=\text{max}(0,\epsilon(q_{\phi}(z|x),A)-\epsilon_0)$, where $\epsilon_0$ is the desired fairness, usually set to $0$ to encourage perfect fairness. Finally, $\mathcal{F}(\phi,A)$ is plugged in Equation \ref{eq:fair_elbo} to jointly optimize $\theta$ and $\phi$ in our fairness-preserving model, which we call the \emph{DF-model}, using the Adam optimization algorithm \cite{kingma2015adam} on the objective via backpropagation \cite{lecun2015deep} and  automatic differentiation \cite{paszke2017automatic}. The pseudo-code to our fair inference approach is provided in the appendix.    
\begin{figure*}[t]
\centering
\resizebox{1.0\textwidth}{!}{
\begin{tabular}{ccc}
	\begin{tikzpicture}
		\node[obs] (x) {$x_{j}^{(D)}$} ;
		\node[latent, above=of x,yshift = -0.5cm, xshift = 1.5cm] (z) {$z_j$} ;
		\node[latent, above=of z, xshift = -1cm, yshift = 0.3cm] (sigZ) {$\sigma_{z}^{(D)}$} ;
		\node[latent, left=of sigZ] (muZ) {$\mu_{z}^{(D)}$} ;
		
		\node[obs, above=of muZ, xshift = 0.5cm, yshift = -0.5cm](sig0){$\sigma_{z}^{(\alpha)}$};
		\node[obs, above=of muZ, xshift = -0.5cm, yshift = -0.5cm](mu0){$\mu_{z}^{(\alpha)}$};
		\node[obs, above=of sigZ,xshift = -0.5cm, yshift = -0.5cm](kappa){$\kappa_{z}^{(\alpha)}$};
		\node[obs, above=of sigZ,xshift = 0.5cm, yshift = -0.5cm](eta){$\eta_{z}^{(\alpha)}$};

 		\edge {mu0} {muZ}
 		\edge {sig0} {muZ}
 		\edge {kappa} {sigZ}
 		\edge {eta} {sigZ}
 		\edge{muZ}{x}
 		\edge{sigZ}{x}
 		\edge{z}{x}

		\plate {param_d} {(muZ)(sigZ)} {$D$}
		\plate {param_z} {(param_d)(mu0)(sig0)(kappa)(eta)} {$z$}
		\plate {data_d} {(x)} {$D$}
		\plate {data_z} {(data_d)(z)} {$j$}
	\end{tikzpicture}
&
	\begin{tikzpicture}
		\node[obs] (x) {$\mathbf{x}_{j}$} ;
		\node[latent, above=of x,yshift = -0.5cm, xshift = 1.5cm] (z) {$z_j$} ;
		\node[latent, above=of z, xshift = -1cm, yshift = 0.3cm] (sigZ) {$\boldsymbol{\Sigma}_{z}^{(D)}$} ;
		\node[latent, left=of sigZ] (muZ) {$\boldsymbol{\mu}_{z}^{(D)}$} ;
		
		\node[obs, above=of muZ, xshift = 0.5cm, yshift = -0.5cm](sig0){$\boldsymbol{\Sigma}_{z}^{(\alpha)}$};
		\node[obs, above=of muZ, xshift = -0.5cm, yshift = -0.5cm](mu0){$\boldsymbol{\mu}_{z}^{(\alpha)}$};
		\node[obs, above=of sigZ,xshift = -0.5cm, yshift = -0.5cm](nu){$\nu_{z}^{(\alpha)}$};
		\node[obs, above=of sigZ,xshift = 0.5cm, yshift = -0.5cm](psi){$\boldsymbol{\psi}_{z}^{(\alpha)}$};

 		\edge {mu0} {muZ}
 		\edge {sig0} {muZ}
 		\edge {nu} {sigZ}
 		\edge {psi} {sigZ}
 		\edge{muZ}{x}
 		\edge{sigZ}{x}
 		\edge{z}{x}

		\plate {param_z} {(param_d)(mu0)(sig0)(nu)(psi)} {$z$}
		\plate {data_z} {(data_d)(z)} {$j$}
	\end{tikzpicture}
&

	\begin{tikzpicture}
		\node[obs] (t) {$t_{j}$} ;
		\node[latent, above=of t, xshift = -2cm, yshift=-0.5cm] (z) {$z_j$} ;
		\node[obs, above=of z, xshift = -0.5cm, yshift=-0.5cm] (f) {$f_j$} ;
		\node[obs, above=of z, xshift = 0.5cm, yshift=-0.5cm] (p) {$p_j$} ;
		\node[obs, above=of z, xshift = -1.5cm, yshift=-0.5cm] (m) {$m_j$} ;
		\node[obs, above=of z, xshift = 1.5cm, yshift=-0.5cm] (d) {$d_j$} ;
	    \node[obs, above=of t, xshift = 0.5cm, yshift=-0.5cm] (a) {$a_j$} ;
	    \node[obs, right=of a, xshift = 1.5cm] (c) {$c_j$} ;
	    \node[latent, above=of a, xshift = 1.8cm, yshift=-0.5cm] (u) {$u_j$} ;
		
		\node[latent, above=of a, xshift = -0.7cm, yshift = 1cm] (muUA) {$\mu_{u}^{(a)}$} ;
		\node[latent, above=of a, xshift = 0.7cm, yshift = 1cm] (sigUA) {$\sigma_{u}^{(a)}$} ;
		\node[latent, above=of c, xshift = -0.7cm, yshift = 1cm] (muUC) {$\mu_{u}^{(c)}$} ;
		\node[latent, above=of c, xshift = 0.7cm, yshift = 1cm] (sigUC) {$\sigma_{u}^{(c)}$} ;
		
		\node[obs, above=of sigUA, yshift=-0.5cm](sigU0){$\sigma_{u}^{(\alpha)}$};
		\node[obs, above=of muUA, yshift=-0.5cm](muU0){$\mu_{u}^{(\alpha)}$};
		\node[obs, above=of muUC, yshift=-0.5cm](kappa){$\kappa_{u}^{(\alpha)}$};
		\node[obs, above=of sigUC, yshift=-0.5cm](eta){$\eta_{u}^{(\alpha)}$};
		
		\node[latent, below=of t, yshift=0.1cm] (betaC) {$\beta_{c}^{(t)}$} ;
		\node[latent, left=of betaC] (betaU) {$\beta_{u}^{(t)}$} ;
		\node[latent, left=of betaU] (betaZ) {$\beta_{z}^{(t)}$} ;
		\node[latent, right=of betaC] (beta0) {$\beta_{0}^{(t)}$} ;
		\node[latent, right=of beta0] (sigZ) {$\sigma^{(t)}$} ; 
		
		\node[obs, below=of betaZ, xshift = -1cm, yshift=0.5cm] (muZ0) {$\mu_{z}^{(\alpha)}$} ;
		\node[obs, right=of muZ0, xshift=-0.8cm] (sigZ0) {$\sigma_{z}^{(\alpha)}$} ;
		\node[obs, right=of sigZ0] (mu0) {$\mu^{(\alpha)}$} ;
		\node[obs, right=of mu0] (sig0) {$\sigma^{(\alpha)}$} ;
		\node[obs, right=of sig0] (k0) {$\kappa^{(\alpha)}$} ;
		\node[obs, right=of k0] (n0) {$\eta^{(\alpha)}$} ;

  		\edge {muU0} {muUA}
  		\edge {muU0} {muUC}
  		\edge {sigU0} {muUA}
  		\edge {sigU0} {muUC}
  		\edge{kappa}{sigUA}
  		\edge{kappa}{sigUC}
  		\edge{eta}{sigUA}
  		\edge{eta}{sigUC}
  		
  		\edge{u}{a}
  		\edge{u}{t}
  		\edge{u}{c}
  		\edge{c}{t}
  		
  		\edge{muUA}{a}
  		\edge{sigUA}{a}
  		\edge{muUC}{c}
  		\edge{sigUC}{c}
  		
  		\edge{m}{z}
  		\edge{f}{z}
  		\edge{p}{z}
  		\edge{d}{z}
  		\edge{z}{t}
  		
  		\edge{betaZ}{t}
  		\edge{betaU}{t}
  		\edge{betaC}{t}
  		\edge{beta0}{t}
  		\edge{sigZ}{t}
  		
  		\edge{muZ0}{betaZ}
  		\edge{sigZ0}{betaZ}
  		\edge{mu0}{betaU}
  		\edge{mu0}{betaC}
  		\edge{mu0}{beta0}
  		\edge{sig0}{betaU}
  		\edge{sig0}{betaC}
  		\edge{sig0}{beta0}
  		\edge{k0}{sigZ}
  		\edge{n0}{sigZ}

 		\plate {param_u} {(muUA)(sigUA)(muUC)(sigUC)(muU0)(sigU0)(kappa)(eta)} {$u$}
        \plate {main_graph} {(m)(f)(p)(d)(z)(u)(a)(c)(t)} {$j$}
        \plate {param_z} {(betaZ)(muZ0)(sigZ0)} {$z$}
        \plate {param_betaU} {(betaU)} {$u$}
        \plate {param_betaC} {(betaC)} {$c$}
	\end{tikzpicture}

	\\
	(a) \textbf{Na\"ive Bayes model} 
&
	(b) \textbf{Gaussian mixture model}
&
    (c) \textbf{Special purpose model}
	
    \end{tabular}
    }
	\caption{Examples of our settings using directed acyclic graphs: (a) Na\"ive Bayes (NB), (b) Gaussian mixture (GMM), and (c) Special purpose (SP) models for $j$ individuals with $D$ attributes. For NB and GMM, $z$ encodes \emph{cluster assignments}. SP model is developed for criminal justice system, where $z$ and $u$ encode \emph{risk of crime} and \emph{systems of oppression}, respectively. Our fair inference technique prevents $z$ from reflecting negative stereotypes based on protected attributes $A$ (not shown in graphs).}
	\label{fig:dags}
\end{figure*}
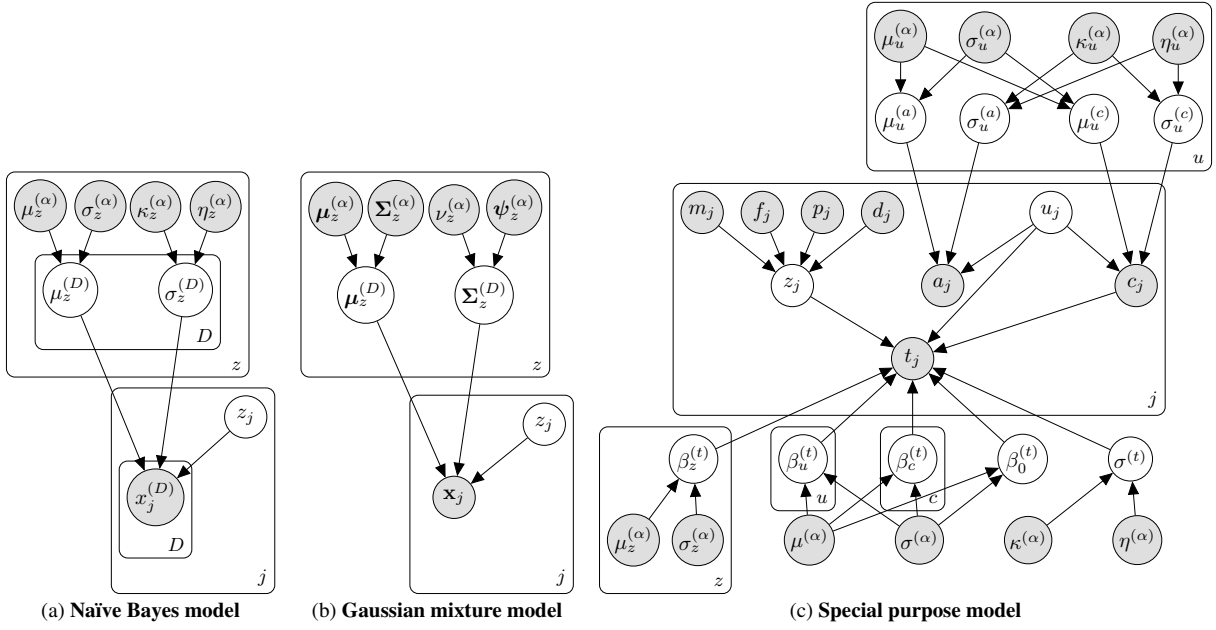

\subsection{Example: Na\"ive Bayes Model}
\label{sec:nb}
We give an example of the na\"ive Bayes (NB) 
model, 
where $x$ is assumed to contain only independent and categorical observed variables. The graphical model is provided in Figure \ref{fig:dags} (a). Let $\theta$ be mean $\mu_{z}^{(D)}$ and standard deviation s.d. $\sigma_{z}^{(D)}$ for a logistic normal which are generated from priors $\mathcal{N}(\mu_{z}^{(\alpha)},\sigma_{z}^{(\alpha)})$ and $\Gamma(\kappa_{z}^{(\alpha)},\eta_{z}^{(\alpha)})$, respectively. Logistic normal is a simpler choice here to reparameterize the model for 
sampling categorical datapoints, as it does not require annealing like Gumbel-Softmax. We can generate samples from the logistic normal as $y_{z}^{(D)}=\mathcal{S}(\mu_{z}^{(D)}+\sigma_{z}^{(D)}\mathcal{E})$, where $\mathcal{S}$ is the softmax function and  $\mathcal{E} \sim \mathcal{N}(0,1)$. Plugging $\log p_{\theta}(x|z) = \sum_{z} \sum_{D} z[x^{(D)}\log y_{z}^{(D)}]$ in Equation \ref{eq:vanilla_elbo} and \ref{eq:fair_elbo} provide \emph{Vanilla-NB} and \emph{DF-NB} models, respectively. 

\subsection{Example: Gaussian Mixture Model}
In our example for Gaussian mixture model (GMM) in Figure \ref{fig:dags} (b), continuous observed variables $\mathbf{x}$ are assumed to be generated from multivariate Gaussian distribution with mean vector $\boldsymbol{\mu}_{z}^{(D)}$ and covariance matrix $\boldsymbol{\Sigma}_{z}^{(D)}$. Let the priors on these parameters be multivariate Gaussian $\mathcal{N}(\boldsymbol{\mu}_{z}^{(\alpha)},\boldsymbol{\Sigma}_{z}^{(\alpha)})$ and inverse Wishart $\mathcal{W}^{-1}(\nu_{z}^{(\alpha)},\boldsymbol{\psi}_{z}^{(\alpha)})$, respectively. However, $\boldsymbol{\Sigma}_{z}^{(D)}$ is a positive semi-definite matrix which is generally impossible to maintain in backpropagation-based gradient methods. To address this, we instead learn real-valued factor $\boldsymbol{\mathcal{C}}_{z}^{(D)}$ of the covariance matrix which is then used to form $\boldsymbol{\Sigma}_{z}^{(D)} = \boldsymbol{\mathcal{C}}_{z}^{(D)} \boldsymbol{\mathcal{C}}_{z}^{(D)^\intercal}+\mathbf{I}$. We then plug $\log p_{\theta}(x|z) = \sum_{z} z[\log \mathcal{N}(\mathbf{x};\boldsymbol{\mu}_{z}^{(D)},\boldsymbol{\Sigma}_{z}^{(D})]$ once again into Equation \ref{eq:vanilla_elbo} and \ref{eq:fair_elbo} to achieve \emph{Vanilla-GMM} and \emph{DF-GMM}, respectively.    

\section{Special Purpose Model for Criminal Justice}
\label{sec:special_purpose}
Our special purpose (SP) model\footnote{
We presented the SP model here in order to demonstrate our fair latent variable modeling and inference methodology. We acknowledge that further investigation and analysis from experts in criminal justice, law and social science would be necessary before considering deployment of our fair SP model in the real systems. After performing such analysis, the eventual goal of this model is that the fairly inferred \emph{risk of crime}, \emph{systems of oppression} and predicted jail time will allow criminal justice professionals to better maintain the right balance between justice, fairness, and public safety. As such, our model represents a small step toward a criminal justice system that is more equitable and fair. 
} is motivated by the ProPublica study \cite{angwin2016machine} on an AI-based risk assessment system called COMPAS, used for bail and sentencing decisions across the U.S. \cite{angwin2016machine} found that COMPAS is almost twice as likely to incorrectly predict re-offending for black people than for white people. 
We develop a special purpose (SP) probabilistic graphical model, that works on top of the system, i.e. COMPAS' predicted score is used as an observation along with regular observed variables which were used to train the COMPAS system, for criminal justice risk assessment using latent variables. The intended use case for the model is to aid judges in bail and sentencing decisions in the court rooms that already use the COMPAS system.  The SP model produces alternate risk scores which aim for fairness by accounting for systemic bias.  We envision that the judges would be presented with both COMPAS risk scores (which emphasize accurate risk assessment but are arguably biased) and the ``fair risk scores'' produced by our system, in order to make balanced and equitable decisions regarding bail and sentencing. We further anticipate that the presence of an alternate ``fair'' risk assessment would encourage judges to think critically about the use of COMPAS, in order to reach a healthy understanding of its limitations. 

In the SP model, we assume that the outcome of the risk assessment mechanism is jail time ($t$) for each offender $j$ which is potentially influenced by some latent variables ($z$ and $u$) and the observed degree of charges ($c$). To design the graphical model in Figure \ref{fig:dags} (c), we look into the existing literature for fairness from diverse fields including AI, humanity, law, and social sciences. Although 
much of the literature in risk assessment views differences in the distributions of risk between protected groups as legitimate phenomena to be accounted for when determining the fairness of a system
\citep{simoiu2017problem}, the intersectionality framework aims for a counterpoint \citep{foulds2020intersectional}. According to intersectionality theory, the distributions of risk are often influenced by unfair societal processes due to systemic structural disadvantages such as racism, sexism, inter-generational poverty, the school-to-prison pipeline, and the prison-industrial complex \citep{collective1977black,crenshaw1989demarginalizing,davis2011prisons,hooks1981ain,wald2003defining}. These unfair processes are termed \emph{systems of oppression}. Inspired by intersectionality framework, we desire to encode a fair and equitable estimate of an individual's \emph{risk of crime} (i.e. low, medium, or high) via $z$ and \emph{systems of oppression} 
via $u$ (here encoded as a binary variable representing the level of impact from these systems), which along with the degrees of charges ($c$), are assumed to affect the jail time ($t$) outcome.  Furthermore, the \emph{risk of crime} ($z$) is considered to be influenced by the offender's historical record, including juvenile misdemeanors ($m$), juvenile felony charges ($f$), previous crime counts ($p$), and the COMPAS system's predicted decile scores ($d$). In contrast, \emph{systems of oppression} ($u$) can lead the structural disadvantages toward the offenders in terms of their age ($a$), degrees of charges ($c$), and jail time ($t$). To reflect these in the graph, we formulate \emph{risk of crime} ($z$) and \emph{systems of oppression} ($u$) as downstream and upstream of the corresponding observed variables, respectively. Since jail time ($t$) is a real-valued observed variable, we formulate it using a regression model with corresponding coefficients $\boldsymbol{\beta}$ for \emph{risk of crime} ($z$), \emph{systems of oppression} ($u$), \emph{degree of charges} ($c$), and an intercept term. We further posit informed hyper-priors on these coefficients to infer identifiable and meaningful latent variables. 


Let 
$p_{\theta}(z|x^{(z)})$ over $z$ be the Gumbel-Softmax whose distribution parameters $\log \pi$ are implemented as neural network outputs, and let prior $p_{\theta}(u)$ over $u$ be the discrete uniform. 
Note that model parameters $\theta$ are weights and biases of $p_{\theta}(z|x^{(z)})$ network, all latent means and standard deviations, and $\boldsymbol{\beta}$. 
From the DAG in Figure \ref{fig:dags} (c), the final objective for our \emph{Vanilla-SP} model is: 
\small
\begin{align}
\label{eq:vanilla_sp}
\mathcal{L}_{\text{SP}}(\theta,\phi) &= \mathbb{E}_{q_{\phi}}[\log p_{\theta}(x^{(z)},x^{(u)},t,z,u)] \nonumber \\&\qquad - \mathbb{E}_{q_{\phi}}[\log q_{\phi}(z,u|x^{(z)},x^{(u)},t)]\nonumber \\
&= \mathbb{E}_{q_{\phi}}[\log p_{\theta}(t|z,u,x^{(u)})+ \log p_{\theta}(x^{(u)}|u) \nonumber\\ 
&\qquad +\log p_{\theta}(z|x^{(z)}) +\log p_{\theta}(u)+p_{\alpha}(\theta)] \nonumber\\
&\qquad - \mathbb{E}_{q_{\phi}}[\log q_{\phi}(z|x^{(z)},t)] - \mathbb{E}_{q_{\phi}}[\log q_{\phi}(u|x^{(u)},t)] \mbox{ ,}
\end{align}

where $\log p_{\theta}(t|z,u,x^{(u)}) = \sum_{z} [\log \mathcal{N}(t;\beta_{0}^{(t)}+z\beta_{z}^{(t)}+\beta_{u}^{(t)}+\beta_{c}^{(t)},\sigma^{(t)})]$ and $\log p_{\theta}(x^{(u)}|u)$ is implemented as log-likelihood of NB model in the previous section. 
Note that $\phi$ represents weights and biases of the inference networks $q_{\phi}(z|x^{(z)},t)$ and $q_{\phi}(u|x^{(u)},t)$. Depending on stakeholders or policy makers, our fairness approach may be applied on any of these variational distributions. In this work, we train \emph{DF-SP} model using Equation \ref{eq:fair_elbo} via Equation \ref{eq:vanilla_sp}, where $\mathcal{F}$ is implemented in terms of $q_{\phi}(z|x^{(z)},t)$, to prevent societal biases in inference on \emph{risk of crime}.    

\section{Practical Considerations for Training}
\label{sec:prac_limits}
Discrete latent variable models are known to be prone to the issue of posterior collapse \cite{bowman2016generating,kingma2016improved,semeniuta2017hybrid,pelsmaeker2020effective}, a particular type of local optimum very close to the prior over latent variables, e.g., all individuals are assigned to same latent class. In practice, we found that better initialization of the parameters and smoothing out the functional space help to resolve this issue, which we implemented by 
using random restarts 
during the hyperparameter grid search on the development (dev) set and by using batch normalization \cite{ioffe2015batch} on the output layer of inference networks, respectively.

However, the above tricks do not resolve the issue in training our SP models. As we optimize a prior network along with model parameters and multiple inference networks, we found that SP models are more prone to posterior collapsing. Existing methods to avoid local optima such as annealing-based approaches to down-weight the KL term \cite{bowman2016generating,alemi2018fixing} in early iterations of the training did not help. 
Finally, we were able to address the problem by using a \emph{warm start} initialization procedure for $p_{\theta}(z|x^{(z)})$ as follows: 1) first pre-train by only maximizing the likelihood $p_{\theta}(t|z)p_{\theta}(z|x^{(z)})$, and 2) then fine tune the prior network, while optimizing the complete ELBO in Equation \ref{eq:vanilla_sp}. 

\section{Experiments}
\label{sec:experiments}
We performed all experiments on the COMPAS dataset\footnote{\url{https://tinyurl.com/2p8tbda2}.} (protected attributes: \emph{race} and \emph{gender}), the Adult 1994 U.S. census income data\footnote{\url{https://archive.ics.uci.edu/ml/datasets/adult}.} (protected attributes: \emph{race}, \emph{gender}, USA vs non-USA \emph{nationality}), and  the Heritage Health Prize (HHP) dataset\footnote{\url{www.kaggle.com/c/hhp}.} (protected attributes: \emph{age} and \emph{gender}). The COMPAS, Adult and HHP datasets contain data instances for a total of 6.91K, 48.84K and 170.07K individuals, respectively. 
Our source code is in the supplementary.

\subsection{Experimental Settings}
\label{sec:exp_settings}
\begin{figure*}[t]
		\centerline{\includegraphics[width=0.86\textwidth]{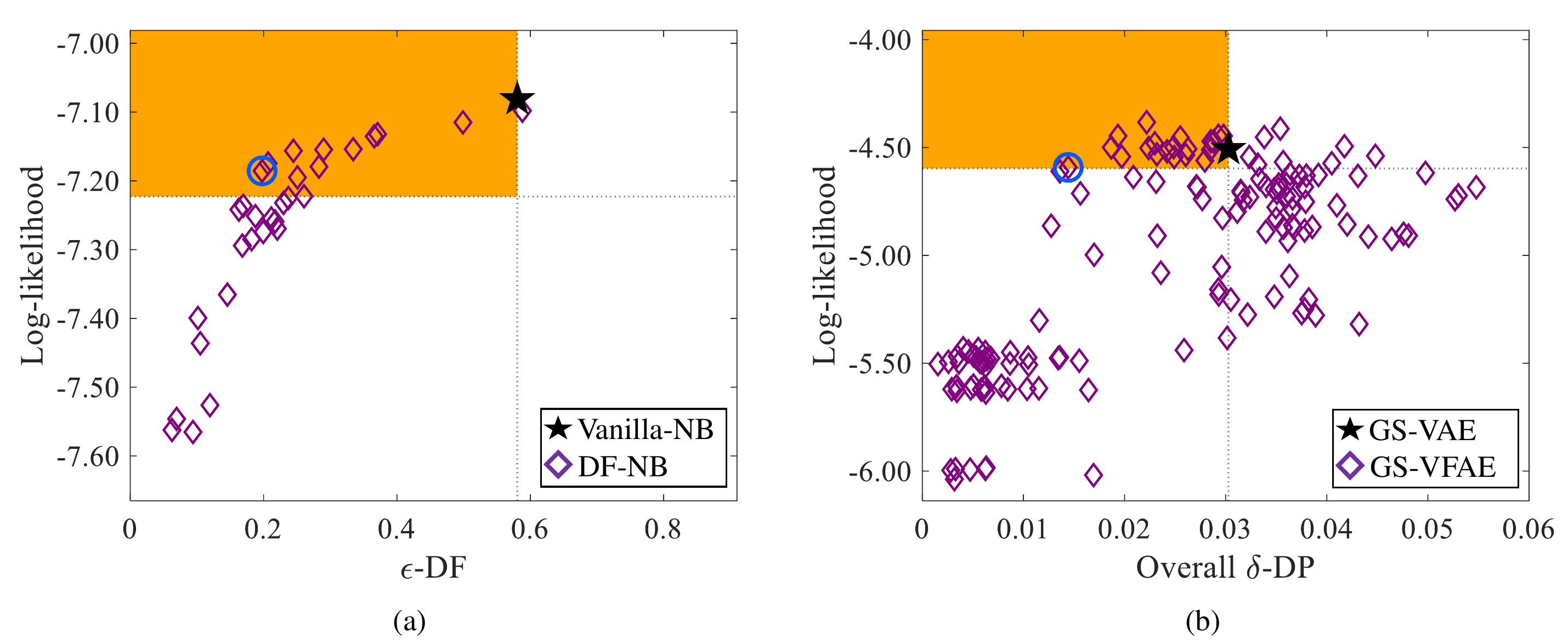}}
		\caption{Selection strategy for fair models (\emph{purple diamonds}): (a) DF-NB and (b) GS-VFAE. \emph{Black asterisk} is the best typical model wrt log-likelihood: (a) Vanilla-NB and (b) GS-VAE. We select the best fair model (\emph{blue circle}) wrt corresponding fairness metric from the \emph{orange area} that satisfies our pre-defined rule.     
		}
		\label{fig:model_select_NB}
\end{figure*}
\begin{table*}[t]
\centering
\resizebox{1.0\textwidth}{!}
{
\begin{tabular}{lccccccccccc}
\toprule
Models & MI $\uparrow$ & CH $\uparrow$ & DB $\downarrow$ & $\epsilon$-DF $\downarrow$ & $\gamma$-SF $\downarrow$ & \begin{tabular}[c]{@{}c@{}}$\delta$-DP $\downarrow$\\ (race)\end{tabular} & \begin{tabular}[c]{@{}c@{}}$\delta$-DP $\downarrow$\\ (gender)\end{tabular} & \begin{tabular}[c]{@{}c@{}}$\delta$-DP $\downarrow$\\ (nation)\end{tabular} & \begin{tabular}[c]{@{}c@{}}$p\%$-Rule $\uparrow$\\ (race)\end{tabular} & \begin{tabular}[c]{@{}c@{}}$p\%$-Rule $\uparrow$\\ (gender)\end{tabular} & \begin{tabular}[c]{@{}c@{}}$p\%$-Rule $\uparrow$\\ (nation)\end{tabular}   \\
\midrule
GS-VAE      & 0.132 & 1391.860 & 3.349 & 0.372  & 0.005 & 0.041 & \textbf{0.005} & 0.011 & 91.452 & \textbf{98.916} & 97.659 \\
GS-VFAE     & 0.087 & 850.065 & 4.286 & 0.329  & \textbf{0.003} & 0.021 & 0.007 & 0.011 & 95.144 & 98.378 & 97.330 \\
Vanilla-NB  & \textbf{0.137} & \textbf{1510.138} & \textbf{2.776} & 0.739  & 0.024 & 0.059 & 0.100 & \textbf{0.004} & 76.080 & 63.130 & \textbf{98.307} \\
DF-NB       & 0.097 & 948.817 & 3.825 & \textbf{0.221}  & 0.005 & \textbf{0.011} & 0.020 & 0.008 & \textbf{96.543} & 93.600 & 97.350 \\
\bottomrule
\end{tabular}
}
\caption{Performance for clustering analysis on categorical variables in Adult dataset. Our DF-NB was the fairest model wrt $\epsilon$-DF along with several other fairness metrics, while our Vanilla-NB performed with highest clustering performances. Higher is better for measures with $\uparrow$, while lower is better for measures with $\downarrow$.}
\label{tab:nb}
\end{table*}
We validate and compare our models with two baseline models. For a typical baseline model that doesn't take fairness into account, we consider the Gumbel-Softmax (GS) reparameterization-based VAE model for discrete latent variables (GS-VAE) \cite{jang2016categorical}. As there is no existing work that enforces fairness in completely unsupervised setting for discrete latent variables, the work from \citet{louizos2016variational} is presumably the most relevant. 
They proposed an unsupervised fair VAE model $p_{\theta}(x|z,A)$ to factor out undesired information from  the continuous latent variables $z$ by the marginally independent protected attributes $A$. We extend this model for discrete latent variables by GS reparameterization and use it as a fair baseline model (GS-VFAE) for our experiments.
 
We split the COMPAS into 60\% train, 20\% dev, and 20\% test sets. For the Adult, we used the pre-specified train (32.56K) and test set (16.28K), and held-out 30\% from the training data as the dev set. Finally, we held-out 10\% from our larger data HHP as the test set, using the remainder for training, and further held-out 10\% from the training data as the dev set. All the models were trained via the Adam optimizer \cite{kingma2015adam} using PyTorch \cite{paszke2019pytorch} on COMPAS, Adult and HHP datasets for a total of 50, 10 and 5 epochs, respectively. Finally. we performed grid search on the dev set to choose hyper-parameters, e.g., minibatch size, \#neurons/hidden layer, learning rate, dropout, activation and random seed, from the same set of hyper-parameter values for all models (see appendix for the set of hyper-parameter values). 
Note that all experiments were conducted using only a CPU on an Intel Xeon E5-2623 V4 server with 64GB memory.

To evaluate the goodness of fit for all models, we compute average log-likelihood (LL) on held-out data. We also compute average mutual information (MI) \cite{kozachenko1987sample,kraskov2004estimating,ross2014mutual} of $z$ over all observed variables to quantify how much meaningful information can be obtained from $z$. For clustering analysis, we measure commonly used metrics like Calinski-Harabasz (CH) score \cite{calinski1974dendrite}, where a higher score represents a model with better defined clusters, and Davies-Bouldin (DB) score \cite{davies1979cluster}, where a lower score represents a model with better separation between the clusters. In our SP model for criminal justice, we evaluate the predictive performance using LL, mean absolute error (MAE), mean squared error (MSE) and regression score ($R^2$) based on observed ``jail time" in held-out data.   

For evaluating from a fairness perspective, we compute $\epsilon$-DF which aims to  ensure equitable treatment for all intersecting groups of all protected attributes. 
We also measure demographic parity ($\delta$-DP) \cite{dwork2012fairness} which ensures similar outcome probabilities for each protected group, $p\%$-Rule \cite{zafar2017fairness} which generalizes the $80\%$ rule of the U.S. employment law \cite{biddle2006adverse} for each protected group,  and subgroup fairness ($\gamma$-SF) \cite{kearns2018preventing} which aims to prevent subset targeting in outcome variable by protecting all specified subgroups. 
To adapt $\gamma$-SF, $\delta$-DP and the $p\%$-Rule to multi-dimensional latent variables, we compute these metrics for each latent class, and report the worst case as the final metric, following \citet{foulds2020intersectional}.

\begin{table*}[t]
\centering
\resizebox{0.85\textwidth}{!}
{
\begin{tabular}{lccccccccc}
\toprule
Models & MI $\uparrow$ & CH $\uparrow$ & DB $\downarrow$ & $\epsilon$-DF $\downarrow$ & $\gamma$-SF $\downarrow$ & \begin{tabular}[c]{@{}c@{}}$\delta$-DP $\downarrow$\\ (age)\end{tabular} & \begin{tabular}[c]{@{}c@{}}$\delta$-DP $\downarrow$\\ (gender)\end{tabular} &  \begin{tabular}[c]{@{}c@{}}$p\%$-Rule $\uparrow$\\ (age)\end{tabular} & \begin{tabular}[c]{@{}c@{}}$p\%$-Rule $\uparrow$\\ (gender)\end{tabular}   \\
\midrule
GS-VAE      & 0.069 & 913.774  & 4.221 & 1.385  & 0.045 & 0.335 & 0.048 & 32.311 & 83.524 \\
GS-VFAE     & 0.061 & 793.340  & 4.523 & 1.123  & 0.042 & 0.320 & \textbf{0.040} & 39.108 & 84.329 \\
Vanilla-GMM & \textbf{0.106} & \textbf{1445.014} & \textbf{2.996} & 2.269  & 0.064 & 0.416 & 0.104 & 16.060 & 77.261 \\
DF-GMM      & 0.044 & 597.302  & 4.183 & \textbf{0.275}  & \textbf{0.021} & \textbf{0.078} & 0.049 & \textbf{84.338} & \textbf{89.964} \\
\bottomrule
\end{tabular}
}
\caption{Performance for clustering analysis on continuous variables in HHP dataset. Our DF-GMM was the fairest model wrt $\epsilon$-DF along with most of the other fairness metrics, while our Vanilla-GMM performed with highest clustering performances. Higher is better for measures with $\uparrow$, while lower is better for measures with $\downarrow$.}
\label{tab:gmm}
\end{table*}
\begin{table*}[t]
\centering
\resizebox{1.0\textwidth}{!}
{
\begin{tabular}{lcccclcccccc}
\toprule
\multirow{2}{*}{Models} & \multicolumn{4}{c}{Measured in terms of observed ``jail time"} &  & \multicolumn{6}{c}{Measured in terms of latent $z$}                                                                                                                                                                                                                                       \\
                        & LL $\uparrow$            & MAE $\downarrow$          & MSE $\downarrow$          & $R^2$ $\uparrow$       &  & $\epsilon$-DF $\downarrow$  & $\gamma$-SF $\downarrow$  & \begin{tabular}[c]{@{}c@{}}$\delta$-DP $\downarrow$ \\ (race)\end{tabular} & \begin{tabular}[c]{@{}c@{}}$\delta$-DP\\ (gender) $\downarrow$ \end{tabular} & \begin{tabular}[c]{@{}c@{}}$p\%$-Rule $\uparrow$\\ (race)\end{tabular} & \begin{tabular}[c]{@{}c@{}}$p\%$-Rule $\uparrow$\\ (gender)\end{tabular} \\ 
\midrule
Vanilla-SP              & \textbf{-1.301}        & \textbf{0.578}        & \textbf{0.757}       & \textbf{0.274}             &  & 1.744         & 0.035       & 0.143                                                        & 0.093                                                          & 40.844                                                      & 43.929                                                        \\
DF-SP                   & -1.358        & 0.662        & 0.926       & 0.112             &  & \textbf{1.304}         & \textbf{0.022}       & \textbf{0.074}                                                        & \textbf{0.048}                                                          & \textbf{41.145}                                                      & \textbf{51.587}                                                        \\ 
\bottomrule
\end{tabular}
}
\caption{Performance for our special purpose model on COMPAS dataset for criminal justice risk assessment. Predictive performances were measured wrt observed ``jail time", while fairness were measured wrt $z$ that encodes \emph{risk of crime}.  Higher is better for measures with $\uparrow$, while lower is better for measures with $\downarrow$.}
\label{tab:sp}
\end{table*}
%
\subsection{Fair Model Selection}
\label{sec:model_select}
Fair models divert the objective from only-the-ELBO to both ELBO and fairness,  which can hurt the predictive performance of the models. Figure \ref{fig:model_select_NB} demonstrates our strategy for a fair model selection on Adult dataset, which we followed for all experiments. We first obtained the best typical model (GS-VAE and our Vanilla-NB) based solely on LL for the dev set via grid search over the same set of hyper-parameter values (\emph{Black asterisk}). Our fair model, DF-NB, was then assigned the same hyper-parameter values as the best Vanilla-NB, while the grid search was conducted over only the fairness trade-off parameter $\lambda$ (\emph{purple diamonds} in Figure \ref{fig:model_select_NB} (a)). As GS-VFAE does not provide any explicit trade-off parameter, we conducted the grid search over all hyper-parameter values (\emph{purple diamonds} in Figure \ref{fig:model_select_NB} (b)) like GS-VAE. Finally, we selected fair models that provide the best corresponding fairness metrics on the dev set, e.g., $\epsilon$-DF for DF-NB  (\emph{blue circle} in Figure \ref{fig:model_select_NB} (a)) and overall $\delta$-DP for GS-VFAE  (\emph{blue circle} in Figure \ref{fig:model_select_NB} (b)), allowing up to a slack tolerance, e.g., 2\% degradation, in LL from the corresponding best typical model (\emph{orange area}). Note that \citet{louizos2016variational} considered $\delta$-DP for a single protected attribute in their VFAE model. Since we consider multiple protected attributes, we selected the best GS-VFAE in terms of an overall $\delta$-DP metric, which is average of $\delta$-DP for each protected attribute. When deploying these methods in practice, the slack tolerance can be amended based on the stakeholders' preferences. 

\subsection{Performance for Clustering}
In this experiment, we evaluated the models on held-out test data for clustering analysis. For Adult data, models were trained on categorical observed variables like work classes, education levels, occupation types and income$\geq$50K or not, where we aimed to infer $z$ that represents whether an individual is ``hard-working" or not. For our NB models, with the prior knowledge on the PDF of a logistic normal distribution, we set priors $\mathcal{N}(2,1)$ and $\mathcal{N}(-2,1)$  on $\mu_z$ to encode ``hard-working'' and not ``hard-working,'' respectively, and the same prior $\Gamma(1,2)$ on $\sigma_z$ for both cases. Table \ref{tab:nb} shows that our Vanilla-NB outperformed all models in terms of clustering performance metrics like MI, CH and DB, while our DF-NB is the best fair model based on $\epsilon$-DF, as well as several other fairness metrics ($\delta$-DP (race) and p\%-Rule race), with a 
small 
cost in 
performance.     

In the HHP dataset, all models were trained on real-valued observations for hospitalized patients like estimation of mortality, drug counts, lab counts and so on, where we aimed to group the patients into 3 clusters that may represent short, medium and long length of stay in hospital so that we can help stakeholders to properly allocate healthcare resources. In our GMM models, we set informed priors on $\boldsymbol{\mu}_z$ using cluster centers from a k-means clustering method on train data and same prior $\mathcal{W}^{-1}(D+2,\mathbf{I})$ on $\boldsymbol{\Sigma}_z$ for all clusters. Table \ref{tab:gmm} shows that our DF-GMM is the fairest model based on almost all fairness metrics (5 out of 6) with a loss in clustering, while our Vanilla-GMM performed as best and worst model with respect to clustering metrics and fairness metrics, respectively.       

\subsection{Performance for Criminal Justice Risk Assessment}
\label{sec:special_purpose_exp}
\begin{table*}[t]
\centering
\resizebox{0.6\textwidth}{!}
{
\begin{tabular}{lcccccc}
\\
\toprule
Models     & $\epsilon$-DF $\downarrow$  & $\gamma$-SF $\downarrow$  & \begin{tabular}[c]{@{}c@{}}$\delta$-DP $\downarrow$ \\ (race)\end{tabular} & \begin{tabular}[c]{@{}c@{}}$\delta$-DP $\downarrow$ \\ (gender)\end{tabular} & \begin{tabular}[c]{@{}c@{}}$p\%$-Rule $\uparrow$\\ (race)\end{tabular} & \begin{tabular}[c]{@{}c@{}}$p\%$-Rule $\uparrow$\\ (gender)\end{tabular} \\ 
\midrule
Vanilla-SP              & \textbf{0.085}         & \textbf{0.005}       & 0.027                                                        & \textbf{0.003}                                                          & 94.323                                                      & \textbf{99.364}                                                        \\
DF-SP                   & 0.100         & 0.006       & \textbf{0.015}                                                        & 0.019                                                          & \textbf{96.873}                                                      & 96.063                                                        \\ 
\bottomrule
\end{tabular}
}
\caption{Fairness metrics measured on latent $u$ which encodes \emph{systems of oppression} against individuals.}
\label{tab:sp_latent_u}
\end{table*}
\begin{figure}[t]
		\centerline{\includegraphics[width=0.6\textwidth]{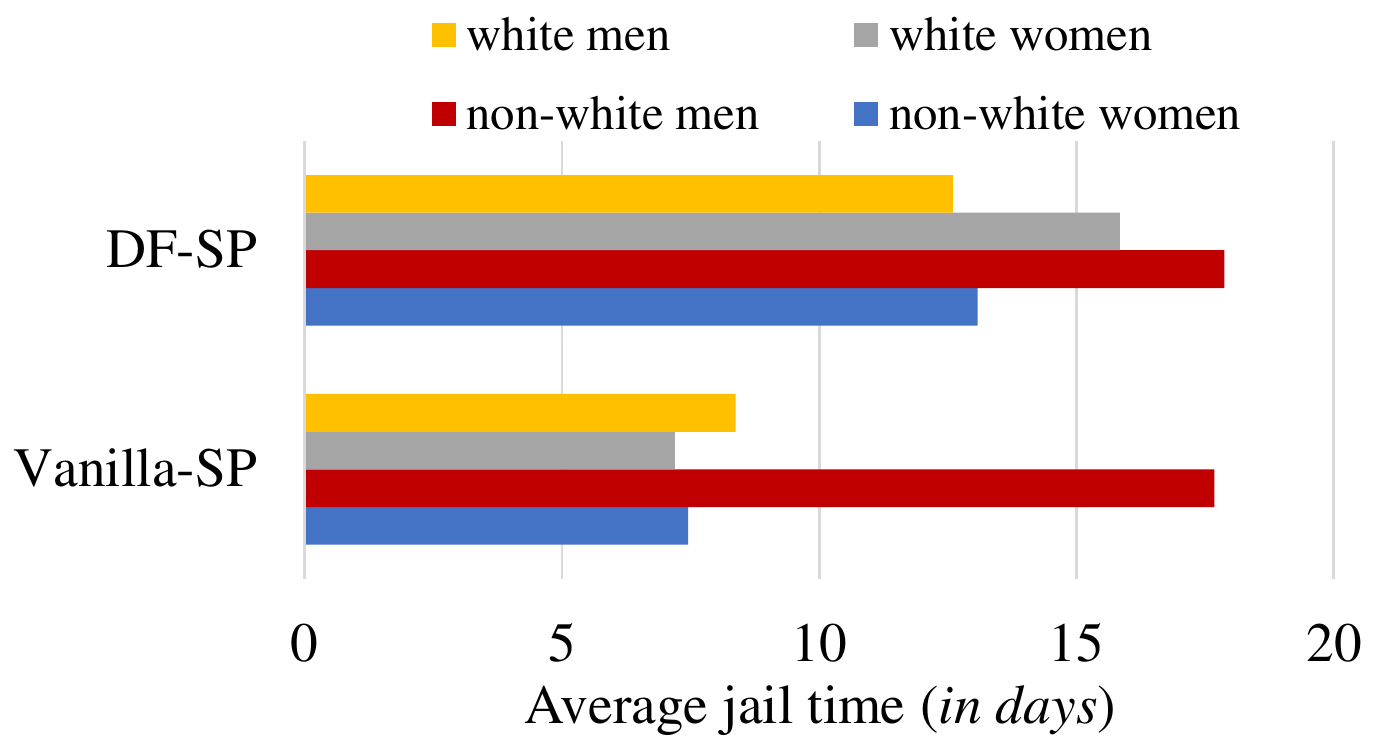}}
		\caption{Generated average ``jail time" in terms of intersecting protected groups for COMPAS data.    
		}
		\label{fig:jailTime}
\end{figure}
%
%
%

We investigate the performance of our SP models on the COMPAS system for criminal justice. 
%
Note that due to the well-known limitations with the COMPAS data \cite{compaslicated21}, we caution against over-interpreting our experimental results. 
In Table \ref{tab:sp}, we show detailed results for our Vanilla-SP and DF-SP models. Note that we excluded the VAE-based baselines from this experiment since there is no straightforward way to extend them for the SP framework. According to the DAG presented in Figure \ref{fig:dags} (c), we can evaluate the overall predictive performance in terms of ``jail time". We measured fairness in terms of latent \emph{risk of crime} $z$ since fairness intervention was applied to $z$. As expected, we found that Vanilla-SP performed better w.r.t predictive performance metrics, but worse w.r.t all fairness metrics. DF-SP  mitigated these biases with a little sacrifice in predictive performances. Table \ref{tab:sp_latent_u} shows fairness measures on latent \emph{systems of oppression} built into our society, where Vanilla-SP outperforms DF-SP model. This result is intuitive from the DAG. Since both \emph{risk of crime} and \emph{systems of oppression} can alter the ``jail time,'' improving fairness for one of them can  increase disparity in the other. 

We also looked into MI for COMPAS's score (MI = 0.079), inferred \emph{risk of crime} by Vanilla-SP (MI = 0.072) and by DF-SP (MI = 0.048) with actually occurred recidivism over a two-year period. Since COMPAS is a supervised learning-based system, it's expected that COMPAS shows higher MI with the actual label, while our unsupervised Vanilla-SP and DF-SP performed with the comparable MI metric. Finally, in Figure \ref{fig:jailTime}, we visualized generated average ``jail time" from our models in terms of all intersecting groups. We observe that Vanilla-SP reflected discrimination by predicting more ``jail time" against a particular group, while DF-SP distributes similar ``jail time" on average for all groups.  

\section{Conclusion}
We have proposed an intersectional fair stochastic inference technique for discrete latent variable models. We have also presented a special-purpose model for 
mitigating societal biases from risk assessments in criminal justice.  Our empirical results show the benefits of our approach in such 
sensitive tasks. %
In future work, we plan to work with criminologists to ensure its proper application and eventual deployment.  We also plan to extend our custom latent variable modeling approach to tackle the case of learning fair algorithms with fully unobserved protected attributes.

\bibliographystyle{unsrtnat}
\bibliography{references}  
\appendix
\section{Related Work}
\label{related_work}
Our fairness intervention technique in this work is inspired by intersectionality, the core theoretical framework underlying the third-wave feminist movement \cite{crenshaw1989demarginalizing,collins2002black}. \citet{foulds2020intersectional} proposed differential fairness which  implements the principles of intersectionality with additional  beneficial  properties  from  a societal perspective regarding the law, privacy, and economics. While most of the fairness notions are defined for binary outcome and binary protected attribute, differential fairness conversely handles multiple outcomes and multiple protected attributes, simultaneously.

Much of the prior work that enforces fairness in variational inference \cite{louizos2016variational,song2019learning,creager2019flexibly,locatello2019fairness} using unsupervised probabilistic graphical models, e.g., VAE \cite{kingma2013auto,rezende2014stochastic}, $\beta$-VAE \cite{higgins2017beta}, and FactorVAE \cite{kim2018disentangling}, aims to learn fair representations of data using continuous latent variables for downstream classification tasks. \citet{louizos2016variational} also proposed a semi-supervised VAE model that encourages statistical independence between continuous latent variables and protected attributes using a maximum mean discrepancy (MMD) \cite{gretton2008kernel} penalty. Through the lens of representation learning, there are other recent advances in building fair classifiers using fair representations. \citet{zemel2013learning} proposed a neural network based supervised clustering model for learning fair representations that maps each data instance to a cluster, while the model ensures that each cluster gets assigned approximately equal proportions of data from each protected group. While this approach cannot leverage the representational power of a distributed representation, other work \cite{edwards2015censoring,madras2018learning,zhao2019inherent,zhao2020conditional} addressed this by developing joint framework using an autoencoder network to learn distributed representations along with an adversary network to penalize when protected attributes are predictable from representations and a classifier network to preserve utility-related information in the representations.  

\section{Pseudo-code for Fair Variational Inference}
\begin{algorithm}[t]
\caption{Intersectional Fair Stochastic Variational Inference}
\label{alg:fairSVI}
\begin{flushleft}
\textbf{Require:} Train data $\mathbf{x}=\{x_{j}\}_{j=1}^{n}$\\
\textbf{Require:} Trade-off parameter $\lambda>0$\\
\textbf{Require:} Desired fairness $\epsilon_0$\\
\textbf{Require:} Constant step-size for expected counts $\rho_{t}$ \\
\textbf{Require:} Constant step-size for optimization algorithm $\rho_{o}$ \\
\textbf{Require:} Randomly initialized generative model's parameters $\theta$, i.e., $\mu$ and $\sigma$ \\
\textbf{Require:} Randomly initialized inference network's parameters $\phi$, i.e., MLP's weights and biases \\
\textbf{Require:} Fixed hyper-priors $p_{\alpha}(\theta)$ \\
\textbf{Require:} Fixed prior $p_{\theta}(z)$\\ 
\textbf{Output:} Likelihood $p_{\theta}(x|z)$\\
\textbf{Output:} Variational Posterior $q_{\phi}(z|x)$\\
\end{flushleft}

\begin{itemize}
\item For each epoch:
    \begin{itemize}
        \item For each minibatch $m$:
        \begin{itemize}
            \item Empirically estimate $\hat{N}_{z,s}=\sum_{x \in \mathbf{x}_{m}: A = s} \mathbf{z}_{m}$ and $\hat{N}_{s}$
            \item Apply update: $N_{z,s} :=(1-\rho_{t})N_{z,s} + \rho_{t}\frac{n}{m}\hat{N}_{z,s}$
            \item Apply update: $N_{s} :=(1-\rho_{t})N_{s} + \rho_{t}\frac{n}{m}\hat{N}_{s}$
            \item Estimate $\epsilon(q_{\phi}(\mathbf{z}|\mathbf{x}),A)$ using $e^{-\epsilon} \leq \frac{N_{z,s_i} + \alpha}{N_{s_i}  + K\alpha}\frac{N_{s_j} + K\alpha}{N_{z,s_j} + \alpha}\leq e^\epsilon$
            \item Compute fairness penalty: $\mathcal{F}(\phi,A)=\text{max}(0,\epsilon(q_{\phi}(\mathbf{z}|\mathbf{x}),A)-\epsilon_0)$
        \item  Apply update using stochastic gradient descent with $\rho_{o}$ via Equation 3 and 4:\\
        \qquad\qquad $\min_{\theta,\phi}\; -\frac{1}{m}\sum_{j=1}^{m}\mathcal{L}_{j}(\theta,\phi) + \lambda \mathcal{F}(\phi,A)$\\
        \qquad\qquad //in practice, \emph{Adam} optimization via backpropagation and  autodiff.
        \end{itemize}
    \end{itemize}
\end{itemize}
\end{algorithm}

To further explain our methodology from Section 3 of the main paper, pseudo-code to our fair inference approach for discrete latent variable models is given in Algorithm~\ref{alg:fairSVI}. 

\section{Hyper-parameter Tuning}
\begin{table*}[]
\caption{Set of hyper-parameter values for the grid search. 
\label{tab:grid_table}}
\centering
\resizebox{0.6\textwidth}{!}
    {
\begin{tabular}{lc}
\toprule
\begin{tabular}[l]{@{}l@{}}\#neurons/hiddens layers \\ for inference network\end{tabular}   & \{{[}64, 64{]}, {[}64, 32{]}, {[}32, 16{]}\} \\
minibatch size                   & \{128, 256\}                                             \\
learning rate                    & \{0.001, 0.002, 0.005\}                                         \\
dropout probability              & \{0.1, 0.25\}                                               \\
activation function              & \{ReLU, SoftPlus\}                                      \\
$l2$ regularization              & \{1$e$-3, 1$e$-4\}                                           \\
$\lambda$ for DF Models                & \begin{tabular}[l]{@{}l@{}}\{0.1, 0.2, 0.5, 0.8, 1.0, 2.0, 3.0, 4.0, 5.0, 6.0, 7.0,\\ 8.0, 9.0, 1.5, 2.5, 3.5, 4.5, 5.5, 6.5, 7.5, 8.5, 9.5,\\ 10.0, 15.0, 20.0, 25.0, 30.0, 50.0, 75.0, 100.0\} \end{tabular}          \\
\bottomrule
\end{tabular}
}
\end{table*}
This additional section complements Section 6.1 in the main paper. 
Table \ref{tab:grid_table} summarizes the set of hyper-parameter values used to perform the grid search on the dev set to choose the best model. First, the best typical models (Vanilla-NB, Vanilla-GMM, Vanilla-SP and GS-VAE) that do not account for fairness were selected based solely on LL for the dev set via grid search over minibatch sizes, \#neurons/hidden layers, learning rates, dropout probabilities and activation functions, while we also considered random restarts to improve initialization with different random seeds for each combination of hyper-parameter values. The DF models (DF-NB, DF-GMM and DF-SP) were assigned the same hyper-parameter values, including selected seed, as the best corresponding vanilla models (Vanilla-NB, Vanilla-GMM and Vanilla-SP, respectively), while the grid search was conducted over only the fairness trade-off parameter $\lambda$. We then selected the best DF model using our pre-defined rule: \emph{select the DF model that provides the best $\epsilon$-DF metric  on the dev set, allowing up to a slack tolerance in LL from the corresponding best vanilla model}. There is no explicit fairness trade-off parameter, e.g., $\lambda$, in GS-VFAE model. Therefore, we conducted a full grid search for GS-VFAE model, similar to the typical models, and then chose the best model using the pre-defined rule: \emph{select the GS-VFAE model that provides the best overall $\delta$-DP metric on the dev set, allowing up to a slack tolerance in LL from the best GS-VAE model}.    
\end{document}